\documentclass[review,11.5pt]{elsarticle}

\usepackage{lineno}
\modulolinenumbers[5]
\usepackage{amsmath,amssymb,amsfonts}
\usepackage{graphicx}
\usepackage{textcomp}
\usepackage{dirtytalk}
\usepackage{adjustbox}
\usepackage[acronym]{glossaries}
\usepackage{multirow}
\usepackage{float}
\usepackage{caption}
\usepackage{subcaption}

\DeclareMathOperator{\Recall}{Recall@N}
\DeclareMathOperator{\True}{True}
\DeclareMathOperator{\avg}{avg}
\DeclareMathOperator{\esc}{esc}
\DeclareMathOperator{\pred}{pred}
\DeclareMathOperator{\train}{train}
\DeclareMathOperator{\test}{test}
\DeclareMathOperator{\val}{val}
\DeclareMathOperator{\RF}{RF}

\usepackage{textcomp}
\def\BibTeX{{\rm B\kern-.05em{\sc i\kern-.025em b}\kern-.08em
    T\kern-.1667em\lower.7ex\hbox{E}\kern-.125emX}}
\usepackage{algorithm} 
\usepackage[noend]{algpseudocode}
\usepackage{etoolbox}
\def\BibTeX{{\rm B\kern-.05em{\sc i\kern-.025em b}\kern-.08em
    T\kern-.1667em\lower.7ex\hbox{E}\kern-.125emX}}
    
\usepackage{hyperref}

\makeatletter
\newcommand*{\algrule}[1][\algorithmicindent]{%
  \makebox[#1][l]{%
    \hspace*{.2em}
    \vrule height .75\baselineskip depth .25\baselineskip
  }
}

\newcount\ALG@printindent@tempcnta
\def\ALG@printindent{%
    \ifnum \theALG@nested>0
    \ifx\ALG@text\ALG@x@notext
    \else
    \unskip
    \ALG@printindent@tempcnta=1
    \loop
    \algrule[\csname ALG@ind@\the\ALG@printindent@tempcnta\endcsname]%
    \advance \ALG@printindent@tempcnta 1
    \ifnum \ALG@printindent@tempcnta<\numexpr\theALG@nested+1\relax
    \repeat
    \fi
    \fi
}
\patchcmd{\ALG@doentity}{\noindent\hskip\ALG@tlm}{\ALG@printindent}{}{\errmessage{failed to patch}}
\patchcmd{\ALG@doentity}{\item[]\nointerlineskip}{}{}{} 
\makeatother

 \journal{}

\begin{document}

\begin{frontmatter}

\title{System Design for a Data-driven and Explainable Customer Sentiment Monitor}

\author[firstaddress,thirdaddress]{An Nguyen} \corref{mycorrespondingauthor}
\cortext[mycorrespondingauthor]{Corresponding authors}
\ead{an.nguyen@fau.de}
\author[secondaddress]{Stefan Foerstel}
\author[fithaddress]{Thomas Kittler}
\author[thirdaddress]{Andrey Kurzyukov}
\author[firstaddress]{Leo Schwinn}
\author[firstaddress]{Dario Zanca}
\author[thirdaddress]{Tobias Hipp}
\author[forthaddress]{Da Jun Sun}
\author[thirdaddress]{Michael Schrapp}
\author[secondaddress]{Eva Rothgang}
\author[firstaddress]{Bjoern Eskofier}
\address[firstaddress]{Machine Learning and Data Analytics Lab, Friedrich-Alexander-Universit\"at Erlangen-N{\"u}rnberg, Germany}

%
\address[secondaddress]{University of Applied Sciences Amberg-Weiden, Germany}
%
\address[thirdaddress]{Siemens Healthcare GmbH, Germany}
%
%
\address[forthaddress]{
Siemens Shanghai Medical Equipment Ltd., China}
%
\address[fithaddress]{infoteam Software AG, Germany}
%
\begin{abstract}
The most important goal of customer services is to keep the customer satisfied. However, service resources are always limited and must be prioritized. Therefore, it is important to identify customers who potentially become unsatisfied and might lead to escalations. Today this prioritization of customers is often done manually.
 Data science on IoT data (esp. log data) for machine health monitoring, as well as analytics on enterprise data for customer relationship management (CRM) have mainly been researched and applied independently.
 In this paper, we present a framework for a data-driven decision support system which combines IoT and enterprise data to model customer sentiment. Such decision support systems can help to prioritize customers and service resources to effectively troubleshoot problems or even avoid them.
The framework is applied in a real-world case study with a major medical device manufacturer. This includes a fully automated and interpretable machine learning pipeline designed to meet the requirements defined with domain experts and end users.
The overall framework is currently deployed, learns and evaluates predictive models from terabytes of IoT and enterprise data to actively monitor the customer sentiment for a fleet of thousands of high-end medical devices. Furthermore, we provide an anonymized industrial benchmark dataset for the research community. 
\end{abstract}
\end{frontmatter}
\newacronym{ml}{ML}{machine learning}
\newacronym{lstm}{LSTM}{long short-term memory}
\newacronym{dl}{DL}{deep learning}
\newacronym{dnn}{DNN}{deep neural network}
\newacronym{nn}{NN}{neural network}
\section{Introduction}
Companies are interested in monitoring the performance of their installed systems. The success of a system depends on the health status of a machine (e.g. derived from IoT data like event logs) and customer perception (e.g. derived from ticket data). However, these two perspectives are mostly separated in the literature. 
%
The machine health perspective is often considered in disciplines like predictive maintenance or more generally prognostic health management \cite{Lei_Li_Guo_Li_Yan_Lin_2018}. Sipos et al. \cite{Sipos_Fradkin_Moerchen_Wang_2014}, for example, used a data-driven approach based on multiple-instance learning from event log data for predictive maintenance for high-end medical devices.
Additionally, event log data are analyzed for intrusion detection \cite{Tuor_Kaplan_Hutchinson_Nichols_Robinson_2017, Kim_Kim_Thu_Kim_2016} or failure detection in data and computing centers \cite{Du_Li_Zheng_Srikumar_2017, Zhang_Xu_Min_Jiang_Pelechrinis_Zhang_2016}.
On the other side, the customer perspective is emphasized in the framework of customer relationship management (CRM), which is a broad discipline including strategies and processes for organizations to handle customer interactions and to keep track of all customer-related information \cite{Soltani_Navimipour_2016}.
Customer escalations are mostly predicted based on ticket data only \cite{Montgomery_Damian_2017, Werner_Li_Damian_2019}.
%
In manufacturing companies (e.g. for medical devices), available data  typically falls in two distinct groups. First is the IoT data/machine logs generated on the system. The second group contains complementary enterprise systems. This includes ticketing systems for service activities, spare part consumption, and reported system malfunctions. 
%
To keep customers satisfied with the operation of their systems is crucial for the success of medical device manufacturer.
Therefore, it is important to identify unsatisfied customers who might lead to escalations.
Hence, a framework making use of both data sources in order to combine these two perspectives would be desirable. Such a system should combine existing IoT (log) data and enterprise data. It could serve as a decision support system for the end user to encourage data-driven and therefore more objective decision-making. 
%
In this paper, we present a fully automated end-to-end machine learning framework which combines both data sources to model customer sentiment. We show that customer sentiment can be better estimated when looking at the system performance based on both the machine log data (e.g. to detect system malfunction affecting the customer) and enterprise data (e.g. ticket data from customer interactions). 
We use historical data of escalations as labels for our predictive models to continuously learn a probability for an escalation as an estimate for the customer sentiment. This resulting decision support system helps to better prioritize customers and trouble shoot problems. 
The concrete problem formulation and proposed solution which combines log and enterprise data to increase predictive power and interpretability for the real-world case study serve as our main contributions. 
The remainder of the paper is structured as follows: Section~\ref{sec:prob_descr} describes the problem to be solved as well as the data sources. 
Section~\ref{sec:meth} describes the overall methodology.
Section~\ref{sec:results} presents the experimental results. Then, Section \ref{sec:discussion} discusses the results and presents the proposed workflow.
Section~\ref{sec:conclusion} concludes our paper and discusses future research directions.

\section{Problem Description}
\label{sec:prob_descr}

\subsection{Business Problem}
Customer satisfaction and hence service resource prioritization is a key priority in many organizations. Here, we analyze data from a large and worldwide installed fleet of high-end medical devices. Therefore, customers, as well as local service entities, naturally differ in the way they communicate and document problems. This inevitably leads to situations where customers facing similar problems address the service provider in vastly different ways.
Hence, objectively prioritizing customers and service resources is a hard problem. Combining relevant information from machine log and enterprise data could potentially help to better understand problems in the field and how they affect the customer sentiment. Therefore, we design a data-driven decision support system to help prioritizing customers based on an estimated sentiment. This can help to minimize unexpected escalations as a product of a more proactive customer support. The case study at hand was conducted with a major medical device company for a fleet of thousands of high-end systems used by customers world wide. 
Major challenges are the amount, heterogeneity, and complexity of the different data sources.

\subsection{Data Description}
In order to solve the business problem at hand we make use of two major data sources which we describe here in more detail.

\subsubsection{Log Data}
\label{sec:logdata}

Log data is a time-based protocol of events recorded by different components of a medical system. An event consists of a timestamp (indicating when the event
occurred), an event source (specifying which system component generated the event), an event id
(representing a category of similar event types by the given event source), an event severity (typically: information, warning and error), and a message text (describing the event and giving more details like sensor values).
Events are defined and implemented by the developers of each particular system component. The severity and amount of sensor data logged is decided by each individual developer. 
\par
Depending on which combination of event-source, event-id and message-text we define as unique, we get approximately $10^5$ different events. There can theoretically be an unlimited number of distinct message texts depending on the usage. One system typically generates from $10^4$ to $10^6$ of these events per day. A typical system family having several thousand installed systems worldwide would then generate  up to $100$ GB of log data per day.
\par
These log files are typically used by customer support centers to diagnose problems as well as by the original system developers to track whether their developed systems work as intended.
\par
Major challenges for analyzing log data are the volume and complexity. We describe later how we automatically extract relevant information from incoming log files.

\subsubsection{Enterprise Data Sources}
\label{sec:enterprisedata}

Enterprise data sources are mostly collected by and stored in enterprise resource planning (ERP) systems. Types of enterprise data are:
\begin{itemize}
    \item service activity data / ticket data - documenting all customer interactions and problems which occur
    \item spare part data - typically related to service activities, includes which spare parts have been used for maintenance / repair of a system
    \item customer base / contract data - listing all customers and the corresponding relationships, especially what kind of service level has been signed
\end{itemize}
Major challenges regarding the enterprise data are:
\begin{itemize}
    \item getting a consistent picture for all customers and service activities worldwide, which is made more difficult by different local ERP systems
    \item manual data inputs contain errors due to typos and incorrect usage
    \item worldwide standards differ a lot, especially since there is no precise definition what a \say{well running medical system} is and, therefore, interpretation of service data can differ from country to country
    \item regional ticket data is often written in the local language
\end{itemize}
Globally operating companies can have several levels of customer service centers ranging from regional to global and all of them are generating ticket data. In our case, we consider three different ticket levels from regional to global. Furthermore, we analyze tickets generated by an information system tracking escalations from customer service to the R\&D department.

\subsection{Requirements}
\label{sec:requirements}

There are special requirements to be met for a successful deployment of a decision support system in a real-world scenario as in the presented case study. We describe these requirements in this section and adapt all design decisions accordingly. During the whole development life cycle from proof-of-concept up to deployment, we worked closely together with domain experts and stakeholders from all relevant departments including potential end users for the implemented decision support system. Thus, we can assure that we meet all requirements and build a framework which has a real impact for the end users. 

\begin{itemize}
   \item \textbf{Dynamics and Efficiency}: Currently, decisions about escalations are made on a weekly basis. Hence, our framework must process data and provide predictions on a weekly basis as well.
   The overall framework should be capable to efficiently load new data, extract features, train a model, and  perform predictions on a weekly basis. This should be done in the time frame of a few hours, e.g. on Monday mornings.
   
   \item\textbf{Model Performance and Output}: The escalation flags (highest escalation level) used as a label in this case study are very sparse and noisy. This causes special challenges for the prediction task. A binary output is not desired, but rather a probability of escalation which models the customer sentiment. Customers with the largest probabilities will then be analyzed in more detail by the end users. Therefore, it is not the main goal to design a machine learning model, which perfectly predicts escalations, but rather a system that helps to identify, based on the designed features, which customers might need special attention.  
   
    \item \textbf{Interpretability}: This is especially important for real-world applications as the present case study since the end user wants to understand the reasoning behind the output of the prediction model not only to take the appropriate actions but also to build trust. We extract specific features from the log and enterprise data. These features were designed together with domain experts and end users to incorporate prior knowledge and interpretable features into the decision support system. 
    
    \item \textbf{Usability}: An interactive application was developed based on a commercial Business Intelligence tool which is currently in use by the medical device provider. Usability also includes considerations of what data sources need to be provided based on the end user's request. The ability to interact with the provided decision support system enables continuous feedback for validation. Based on the explanations of the model and features, the end user can decide if the predictions are valid and with that provide more and cleaner labels for future prediction cycles. We will later describe an envisioned workflow for the designed decision support system.
\end{itemize}

\section{Methodology}
\label{sec:meth}
\begin{figure}[H]
  \centering
  \includegraphics[width=\linewidth]{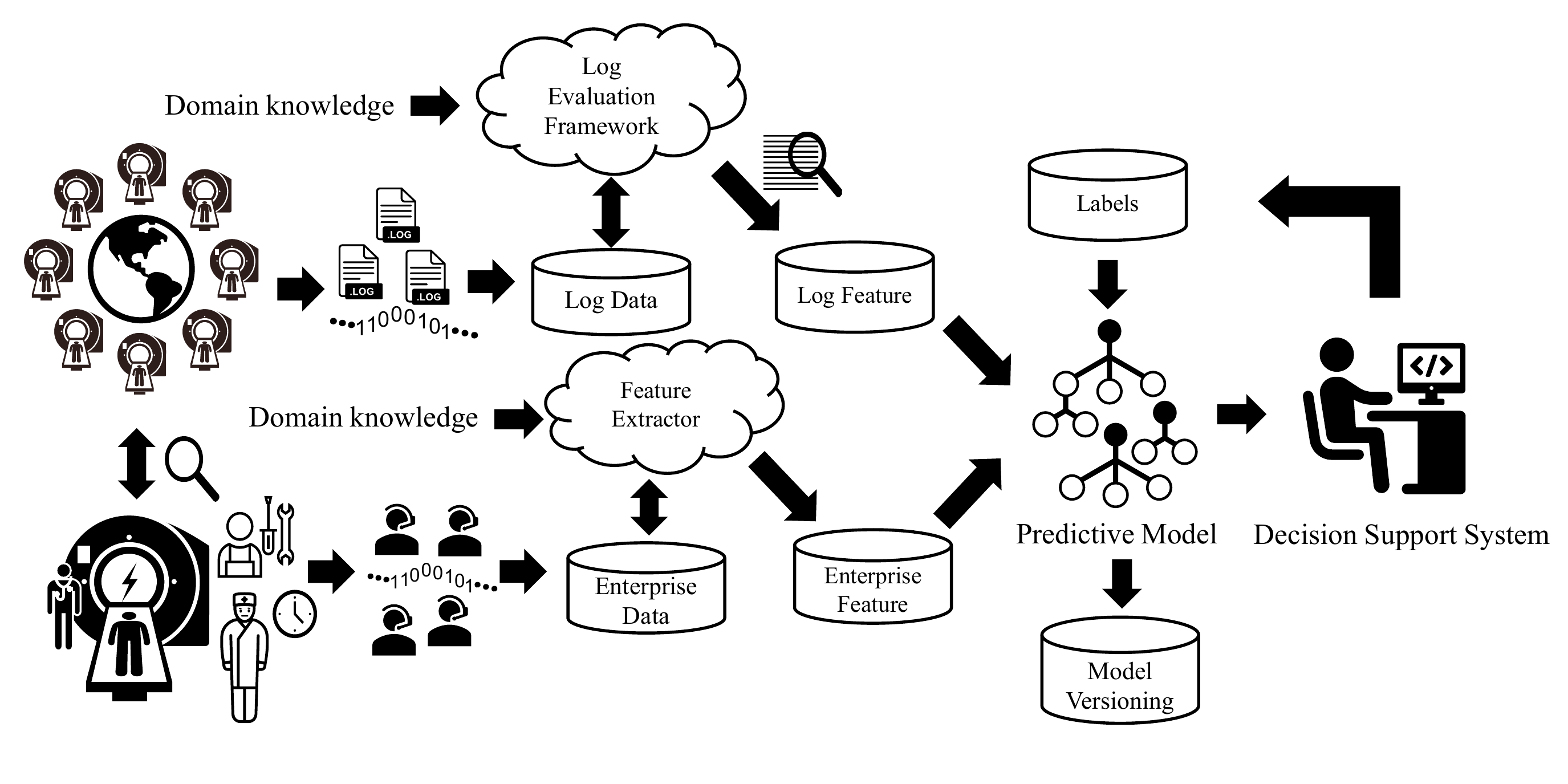}
  \caption{High-level schema of the overall processing pipeline. Systems from all over the world are sending log data. Additionally, enterprise data (sales and ticket data) are collected. Features are extracted based on domain knowledge in order to train a predictive model. The resulting data and predictions are integrated into a decision support system for the end user.}
  \label{fig:overview}
\end{figure}

In this section, we describe the designed framework to solve the business problem at hand. A high-level overview of the implemented framework is depicted in Fig. \ref{fig:overview}. Hundreds of gigabytes of incoming log and enterprise data from all over the world are automatically analyzed via a log evaluation framework to calculate relevant features designed together with domain experts. We design an automated and interpretable machine learning pipeline to calculate a probability of escalation, which models the customer sentiment. The provided decision support system includes an explanation for the calculated probability, as well as historical feature data based on extracted log and enterprise data for the end user. There are major benefits when combining both data sources from the user perspective. Depending on which features explain a prediction, it is possible to identify problems as either being more related to R\&D (log features) or customer service (enterprise features). 

%
%
%
\subsection{Build Dataset}
Algorithm \ref{alg:build_dataset} describes how we built the dataset for the experimental setup. In the following, we will describe the process in more detail. 
\subsubsection{Labeling}
\label{sec:labeling}
\begin{figure}[h!]
  \centering
  \includegraphics[scale=0.4]{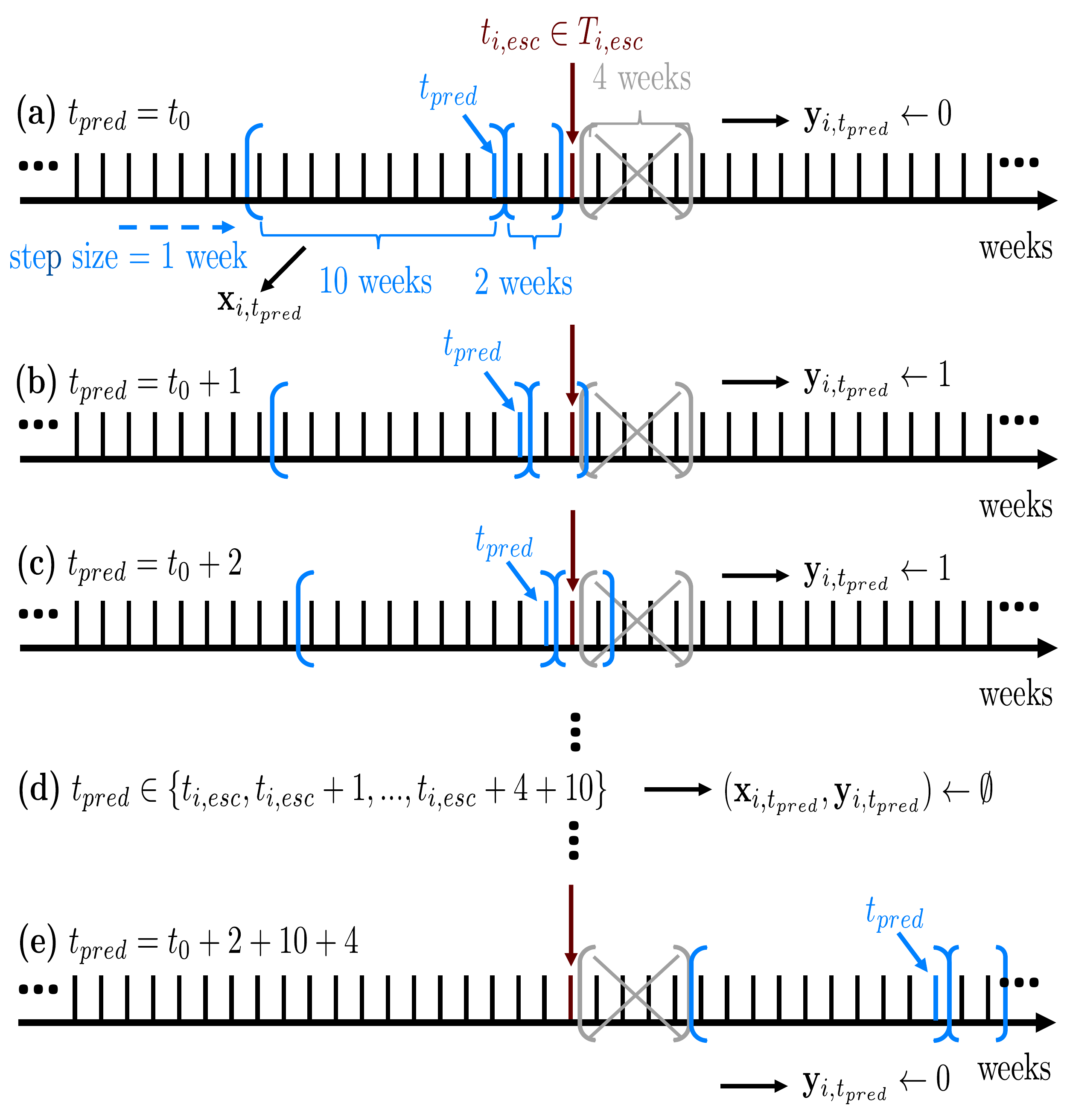}
  \caption{Description of the labeling approach for one example customer $i \in I$. We assume that $t_{\pred}$ (last week in window) starts at some point $t_0$. \textbf{(a)} Sample with negative label ($\mathbf{y}_{i, t_{\pred}} \gets 0$
) since there is no escalation within predictive interval of $2$ weeks. \textbf{(b)}, \textbf{(c)} Samples are labeled positive 
($\mathbf{y}_{i, t_{\pred}} \gets 1$)
since there is an escalation within the predictive interval. \textbf{(d)} We exclude infected samples from the dataset. \textbf{(e)} First valid sample after the infected interval.}
  \label{fig:labeling}
\end{figure}

\begin{figure}[h!]
  \centering
  \includegraphics[width=0.9\linewidth]{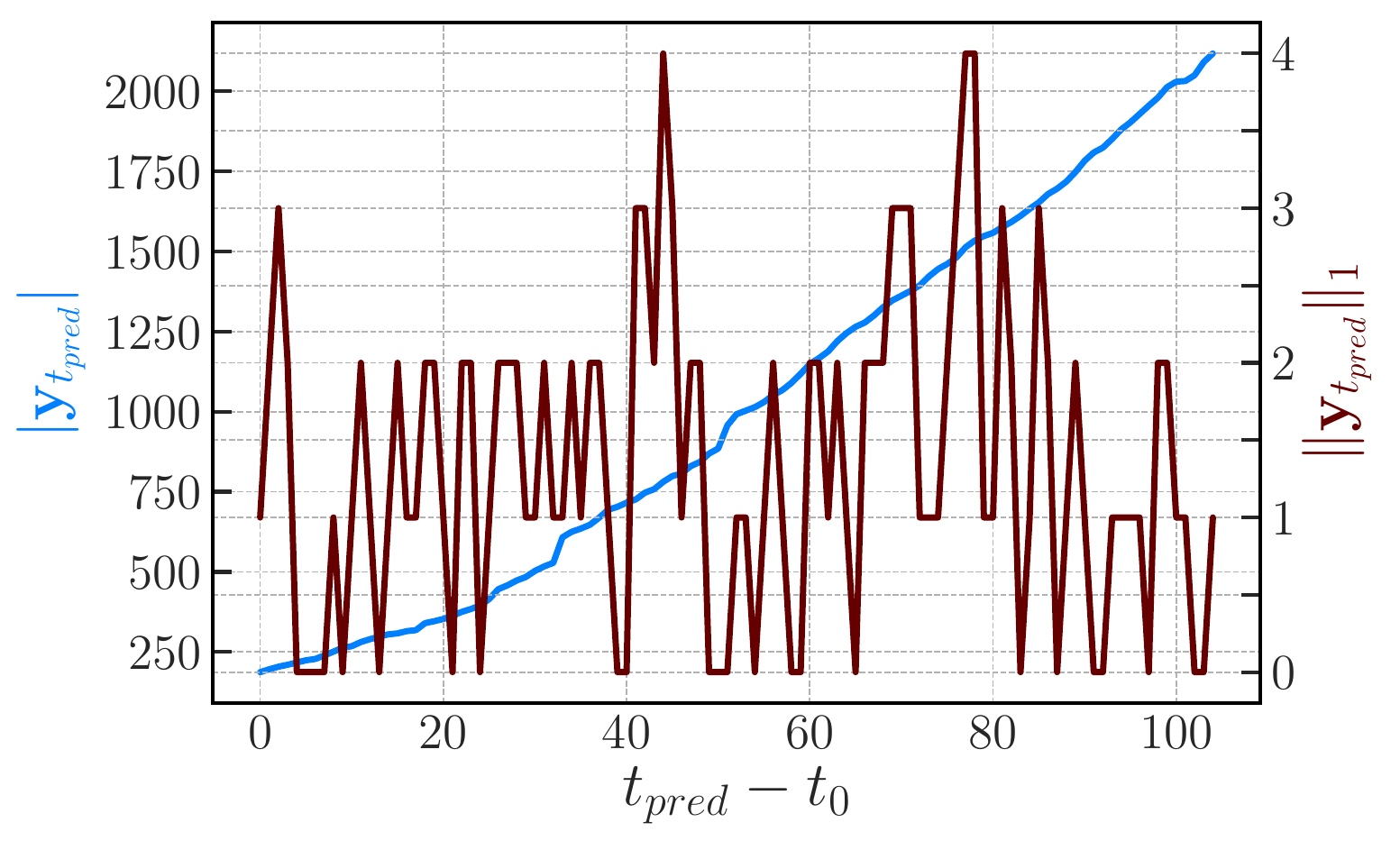}
  \caption{Sample distribution for the whole dataset $\mathcal{D}$. The graph depicts all samples ($|\mathbf{y}_{t_{\pred}}|$) and positive samples ($\|\mathbf{y}_{t_{\pred}}\|_1$) for $t_{\pred} = t_0 + \{0, \ldots, 104\}$.}
  \label{fig:sample_distr}
\end{figure}

Let $I$ be the set of all customers.
Fig. \ref{fig:labeling} depicts the labeling approach for one example customer $i \in I$ using a sliding window approach. The step size is set to $1$ week. We set the window size to $10$ steps, which was proposed by domain experts. Different values were also evaluated but did not yield an improvement. From this window a feature vector $\mathbf{x}_{i, t_{\pred}}$ is extracted, where $t_{\pred}$ is defined as the last week in a window.
Let $T_{i, \esc}$ be the set of all escalations flags for customer $i$ and $t_{i, \esc}$ a specific time point of an escalation flag for customer $i$.
The predictive interval is set to $2$ steps. If there has been an escalation $t_{i, \esc} \in T_{i, \esc}$ in the predictive interval, the label ($\mathbf{y}_{i, t_{\pred}}$) for the sample will be set to $1$ and $0$ otherwise (line \ref{l9}-\ref{l12}). 
After an escalation $t_{i, \esc}$, the following $4$ steps are defined as an \textit{infected interval}. All samples where the sliding window contains weeks from the infected interval are excluded (line \ref{l14}-\ref{l15}). This was defined together with domain experts. We assume that there is already a special focus on customers for which a recent escalation occurred.
As described in line \ref{l2}-\ref{l3}, we repeat this procedure for all customers for a fixed time frame of $104$ steps, which in our case is equivalent to $2$ years.
The dataset $\mathcal{D}_{t_{\pred}} = (\mathbf{x}_{t_{\pred}}, \mathbf{y}_{t_{\pred}})$ contains samples for all customers with complete data (line \ref{l4}).
This results in the distributions depicted in Fig. \ref{fig:sample_distr}. Note that the number of customers $|\mathbf{y}_{t_{\pred}}|$ is increasing while the number of escalations $\|\mathbf{y}_{t_{pred}}\|_1$ remains almost constant over time. This is due to limited service resources which yields to an approximate constant number of customers put into focus each week.
We provide an anonymized version of $\mathcal{D}$ for the research community as an industrial benchmark \cite{nguyen_an_2020_4383145} \footnote{\url{https://doi.org/10.5281/zenodo.4383145}}.
\subsubsection{Feature Extraction}
\label{sec:features}
Weekends and public holidays can introduce noise into calculated features. Therefore, we decided together with domain experts to aggregate features on a weekly basis. For all customers $i \in I$ and prediction weeks $t_{\pred}$, we calculate features for a window of $10$ weeks (line \ref{l2}-\ref{l8}).
%
\\
\\
\noindent
\textbf{Log Data}\\
Due to the high volume and complexity of existing data sources, feature extraction is required. The machine log data as described in \ref{sec:logdata} is not feasible to analyze in their raw format. We use a log evaluation framework to detect the occurrences of specific sequences of events, determined by domain experts. The extracted features have clear meanings and are related to specific system malfunctions, which can affect customers in their daily work routines. Such features include, for example: abort of operation, system delay, user interface (UI) freeze, and UI pop ups. We also extract whether there was a software (SW) update performed for a system.
\\
\\
\textbf{Enterprise Data}\\
The enterprise data available can be split in two connected groups - sales data and customer service tickets - as described in \ref{sec:enterprisedata}. 
Features for sales data are the number and total cost of replaced parts.
Features derived from ticket data include the number of open tickets, age of the oldest open ticket, the rated severity, as well as the number of site visits for each customer depending on availability in the different ticketing systems. These features can be extracted on a global level.
\subsection{Modeling}
In order to meet the requirements described in Section \ref{sec:requirements}, we selected a specific approach to model the customer sentiment. Major challenges are the large class imbalance (Fig. \ref{fig:sample_distr}) and significant amount of label noise as a result of manual decision for an escalation. The output of a machine learning model can be helpful in several ways. First, we can identify which specific problems depicted in the designed features lead to escalation. Additionally, we can identify customers which have similar problems and might need special attention. We compare different machine learning methods: ensembles of decision trees and \glspl{dnn}. For both methods, we calculate post-hoc explanations for each prediction using either a tree explainer \cite{Lundberg_Erion_Chen_DeGrave_Prutkin_Nair_Katz_Himmelfarb_Bansal_Lee_2020} or a modification of DeepLift \cite{Lundberg_Lee_2017}. Both algorithms are implemented in the SHAP libary \cite{Lundberg_Lee_2017}\footnote{\url{https://github.com/slundberg/shap}} and we refer to the explanatory outputs as SHAP values.
\subsubsection{Ensemble of Decision Trees}
Ensemble of decision tree methods have the following benefits:
\begin{itemize}
    \item The computed feature importance \cite{Genuer_Poggi_Tuleau-Malot_2010, Hastie_Tibshirani_Friedman} helps end users understand which of the designed features are \say{correlating} with escalations/customer sentiment.
    \item Ensemble methods provide a probability as a model output which can be interpreted as the customer sentiment (probability for escalation).
    \item Since each combination of time point (week) in a window and designed feature is modeled as a single input variable, we can provide the relevance of each input variable for all predictions to the end user for better troubleshooting.
\end{itemize}
 The decision tree ensemble methods we select are Random Forest (RF) \cite{breiman2001}\footnote{\url{https://scikit-learn.org/stable/modules/generated/sklearn.ensemble.RandomForestClassifier.html}} and XGBoost (XGB) \cite{Chen_Guestrin_2016} \footnote{\url{https://xgboost.readthedocs.io/en/latest/python/python_api.html}}.
Random Forest and XGBoost are ensemble learning techniques which can be used for both classification and regression. In our case, we are interested in classification. 
In general, a $\RF$ is a collection of weak classifiers $\{C_i\}$ where each classifier gets the same input $\mathbf{x}$ and outputs the most probable class $C_i(\mathbf{x}) = y_i \in S$ with $S$ being the set of all possible classes. The output of the Random Forest is then defined as
\begin{equation*}
    \RF(\mathbf{x}) = {\arg\max} |\{y_i : y_i = C_i(\mathbf{x})\}| = y^* \in S,
\end{equation*}
the class which is most probable for the majority of the $C_i$. In the original paper\cite{breiman2001} and also in our case, we used decision trees as weak classifiers.
The decision trees for RF are generated independently and in parallel via a bagging (bootstrap aggregation) approach. This means that each decision tree is generated in two steps:
\begin{enumerate}
    \item Bootstrapping: Independently sampling the input data set $\mathcal{D}_{train} = {(\mathbf{x}_j, y_j)}$ with $j \in \{1, \ldots, m\}$ for each $C_i$ on data points and features. \\
    This means the data points from $\mathcal{D}$ are sampled iid (independent and identically distributed) into a subset $\mathcal{D}_{train_i} = {(\mathbf{x}^i_j, y^i_j)}_{j \in J_i}$ where $J_i \subset \{1, \ldots, m\}$. \\
    Also, the feature space is sampled iid, such that if $\mathbf{x_j}$ contains the features $\mathcal{F} = \{f_k : k \in \{1, \ldots, n\}\}$, then $\mathbf{x}^i_j$ contains the features from $F^i \subset \mathcal{F}$.
    \item Aggregating: Averaging or in our case deciding by a majority vote which class should be chosen.
\end{enumerate}
Gradient Boosting\cite{Friedman2001} also combines many weak classifiers into a strong classifier, but the idea how to combine those weak learners differs. In contrast to bagging, the decision trees are not built in parallel but sequentially, while results are combined along the way. In our case, we chose XGBoost \cite{Chen_Guestrin_2016} which is an improved variant of the Gradient Boosting algorithm using a more regularized formalization of the model leading to a reduction of over-fitting.
In both cases the output is the predicted class (here: $0$ or $1$) as well as the probability the model assigns to each prediction. We use the probability the model assigns to class $1$ as the predicted customer sentiment $\mathbf{\hat{y}}_{t_{\pred}}$ for each input $\mathbf{x}_{t_{\pred}}$.
We address the imbalanced class problem by applying either random oversampling of the minority class, SMOTE \cite{Chawla_Bowyer_Hall_Kegelmeyer_2002} or random undersampling of the majority class \cite{He_Garcia_2009} using the imblearn libary \cite{imblearn2017}\footnote{\url{https://github.com/scikit-learn-contrib/imbalanced-learn}}. We treat the sampling strategy as a hyperparameter in our model selection approach, which we will describe later.
We tested 8 different model configurations as summarized in Table \ref{tab:exp}. We applied two different data fusion approaches. For \say{early} fusion (M1, M2) we simply stack enterprise ( $\mathbf{x}_{ent}$) and log ( $\mathbf{x}_{log}$) features to train a single classifier (RF or XGB). In \say{late} fusion (M3, M4) we train one base classifier based on $\mathbf{x}_{ent}$ and one based on $\mathbf{x}_{log}$. The output of each base classifier is then fed into a subsequent logistic regression layer for the final prediction. Both base classifier are either RF or XGB. We additionally tried to train a classifier only based on $\mathbf{x}_{ent}$ (M5, M6) or only on $\mathbf{x}_{log}$ (M7, M8).
\\
\subsubsection{Deep Neural Networks}
\begin{figure}[h!]
  \centering
  \includegraphics[width=0.3\linewidth]{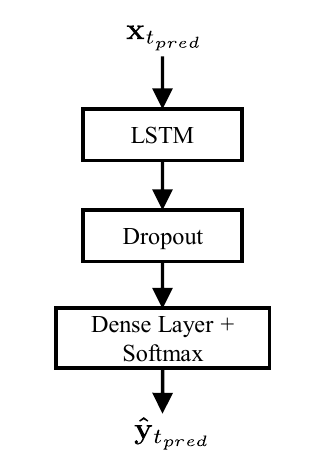}
  \caption{Implemented \gls{dnn} architecture.}
  \label{fig:DL}
\end{figure}

 We implemented a \gls{dnn} based on \gls{lstm} \glspl{nn} \cite{hochreiter.1997} in order to model $x_{t_{\pred}}$ as a time series. 
 Given a sequence of inputs $x_{t_{\pred}}=\langle\mathbf{x}^{(1)}, \mathbf{x}^{(2)},  \ldots, \mathbf{x}^{(10)}\rangle$, a LSTM computes sequences of outputs
$\langle\mathbf{h}^{(1)}, \mathbf{h}^{(2)}, \ldots, \mathbf{h}^{(10)}\rangle$ via the following recurrent equations:
 \begin{align*}
\label{eq:lstm}
    \mathbf{f}_{g}^{(t)}&=\sigma(\mathbf{W}_{f}[\mathbf{x}^{(t)}, \mathbf{h}^{(t-1)}]+\mathbf{b}_{f}) & \text{(forget gate)}, \nonumber \\ 
    \mathbf{i}_{g}^{(t)}&= \sigma(\mathbf{W}_{i}[\mathbf{x}^{(t)}, \mathbf{h}^{(t-1)}]+\mathbf{b}_{i}) & \text{(input gate)}, \nonumber \\
    \mathbf{\Tilde{c}}^{(t)}&= \tanh(\mathbf{W}_{g} [\mathbf{x}^{(t)}, \mathbf{h}^{(t-1)}]+\mathbf{b}_{g}) & \text{(candidate memory)}, \nonumber \\
    \mathbf{c}^{(t)}&= \mathbf{f}_{g}^{(t)}  \circ \mathbf{c}^{(t-1)}+\mathbf{i}_{g}^{(t)} \circ \mathbf{\Tilde{c}}^{(t)} & \text{(current memory)}, \\ 
    \mathbf{o}_{g}^{(t)}&= \sigma(\mathbf{W}_{o} [\mathbf{x}^{(t)}, \mathbf{h}^{(t-1)}]+\mathbf{b}_{o}) & \text{(output gate)}, \nonumber \\
    \mathbf{h}^{(t)}&= \mathbf{o}_{g}^{(t)} \circ \tanh(\mathbf{c}^{(t)}) & \text{(current hidden state)}, \nonumber \\
    &\forall t \in \{1,2,\dots, 10\}. \nonumber
    \end{align*}
$\{\mathbf{W}_{f, i, g, o}, \mathbf{b}_{f, i, g, o}\}$ are trainable parameters, $\sigma$ is the sigmoid activation function,  $\circ$ denotes the Hadamard product (element-wise product), $\mathbf{h}^{(t)}$ and $\mathbf{c}^{(t)}$ are the hidden state and cell memory of a LSTM cell. 
Additionally, a LSTM cell uses four gates to manage its states over time to avoid the problem of exploding/vanishing gradients in the case of longer sequences~\cite{bengio.1994}.
$\mathbf{f}_{g}^{(t)}$ (forget gate) determines how much of the previous memory is kept, $\mathbf{i}_{g}^{(t)}$ (input gate) controls the amount new information ($\mathbf{\Tilde{c}^{(t)}}$) stored into memory, and $\mathbf{o}_{g}^{(t)}$ (output gate) determines how much information is read out of the memory. The hidden state $\mathbf{h}^{(t)}$ is commonly forwarded to a successive layer. In our experiments, we set the number of LSTM layers to $\in \{1, 2\}$ as a hyperparameter. Additionally, we set as a hyperparameter if the LSTM is bidirectional \cite{Schuster_Paliwal_1997} or not. The final output vector from the last LSTM cell $\mathbf{h}^{10}$ is then forwarded to a fully connected layer using dropout \cite{srivastava2014dropout} for regularization and softmax activation for prediction. Fig. \ref{fig:DL} depicts the implemented \gls{dnn} architecture. We address the imbalanced class problem by applying either random oversampling of the minority class or random undersampling of the majority class \cite{He_Garcia_2009}. We used the PyTorch framework \footnote{\url{https://pytorch.org/}} to implement our \gls{dnn} architecture. 
\subsection{Training and Validation}
\begin{figure}[h!]
  \centering
  \includegraphics[width=0.95\linewidth]{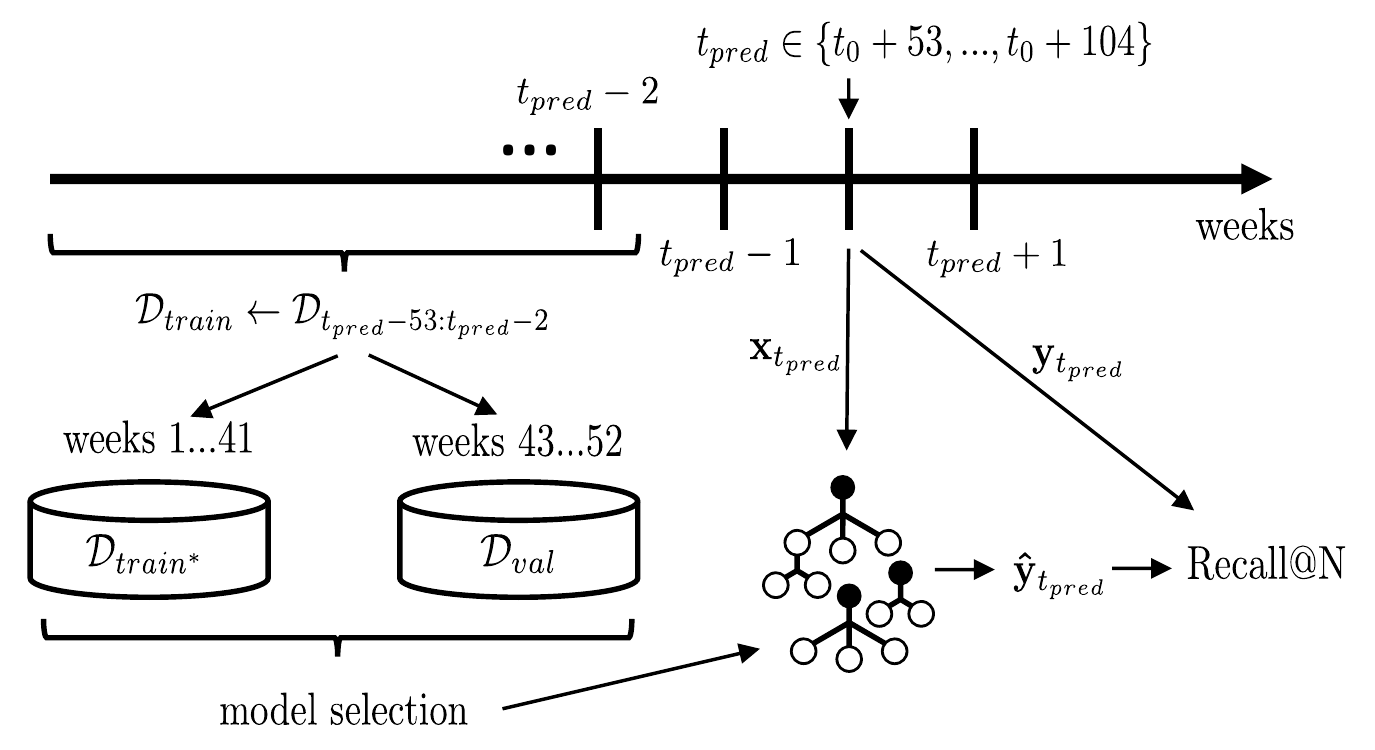}
  \caption{Illustration of the proposed training and evaluation setup (Algorithm \ref{alg:train_val}). We evaluate the decision support system on a weekly basis for one year ($t_{\pred} \in \{t_0+53, \ldots, t_0+104 \}$). For each week we use the data from the previous year ($52$ weeks) to train a model and to get a probability output for each customer ($\mathbf{\hat{y}}_{t_{\pred}}$). Finally, we calculate $\Recall$ to evaluate the performance over the whole year.}
  \label{fig:train_val}
\end{figure}

Our experiments are designed to simulate the real-world performance of our decision support system. Algorithm \ref{alg:train_val} line \ref{l16}-\ref{l24} and Fig. \ref{fig:train_val} depict the experimental setup.
\subsubsection{Evaluation Metrics}
For a given set of estimated customer sentiment values $\mathbf{\hat{y}}_{t_{\pred}}$ and ground truth escalation labels $\mathbf{y}_{t_{\pred}}$, we calculate the $\Recall$ for a whole year $t_{\pred} = t_0 + \{53, \ldots, 104\}$ as an evaluation metric. $\|\mathbf{y}_{t_{\pred}}\|_1$ is the number of positive samples  and $|\mathbf{y}_{t_{\pred}}|$ the total number of samples at $t_{\pred}$, respectively. We define
\begin{equation*}
    \Recall = 100\cdot\frac{\sum_{t_{\pred}} \True(N(\mathbf{\hat{y}}_{t_{\pred}}))}{\sum_{t_{\pred}}\|\mathbf{y}_{t_{\pred}}\|_1},
\end{equation*}
where $N()$ denotes the $N$ largest elements and $\True()$ denotes the number of samples which have a positive ground truth value. Furthermore, we define 
\begin{equation*}
    \avg(\Recall) = \frac{\sum_{N=1}^{100}\Recall}{100},
\end{equation*}
as the average $\Recall$, in order to compare different models for a relevant range of values $N$.
\subsubsection{Model Selection and Evaluation}
\begin{figure}[h!]
     \centering
     \includegraphics[width=0.9\linewidth]{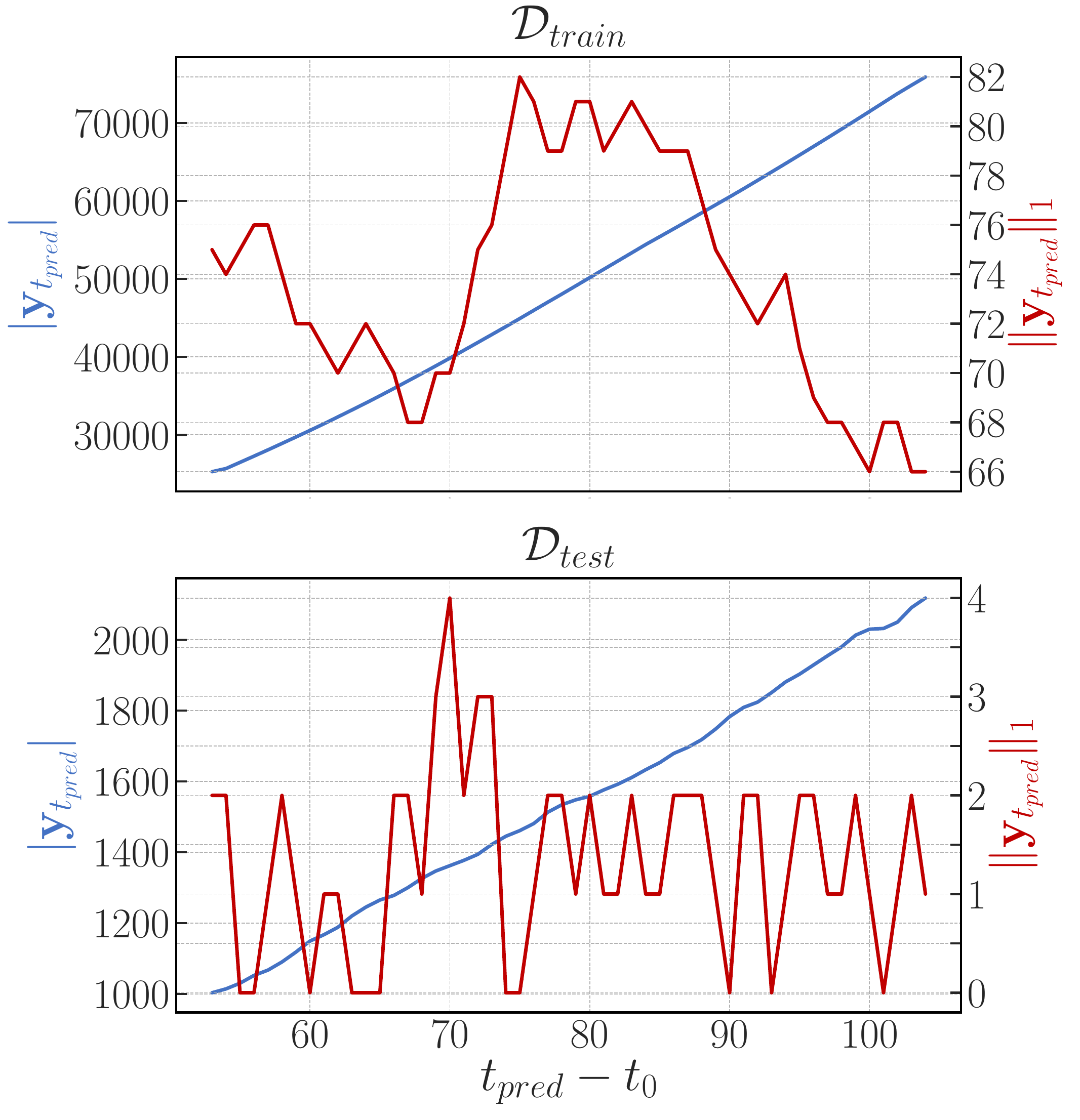}
    \caption{Resulting distribution of samples in $\mathcal{D}_{train}$ and $\mathcal{D}_{test}$. The graphs depict all samples ($|\mathbf{y}_{t_{\pred}}|$) and positive samples ($\|\mathbf{y}_{t_{\pred}}\|_1$) for $t_{\pred} = t_0 + \{53, \ldots, 104\}$.}
    \label{fig:distr_train_test}
\end{figure}

 We perform a weekly analysis for one year ($t_{\pred} = t_0 + \{53, \ldots, 104\}$) to evaluate our approach. For each week, we use the past year as training data ($\mathcal{D}_{\train} \gets \mathcal{D}_{t_{\pred}-53:t_{\pred}-2}$). The gap of $2$ weeks are needed since in deployment we would have complete data until $t_{\pred}$. Therefore, we only know the label for samples until $t_{\pred}-2$, given our predictive interval is $2$ weeks. Fig. \ref{fig:distr_train_test} shows the resulting distributions for $\mathcal{D}_{train}$ and $\mathcal{D}_{\test}$.
 We additionally split the training data for model selection into the first $41$ ($\mathcal{D}_{\train^*} \gets \mathcal{D}_{t_{\pred}-53:t_{\pred}-13}$) and last $10$ ($\mathcal{D}_{\val} \gets \mathcal{D}_{t_{\pred}-11:t_{\pred}-2}$) weeks. We apply the tree-structured Parzen estimator (TPE) \cite{Bergstra_2011} approach for hyperparameter tuning. TPE is a Bayesian optimization approach for hyperparameter tuning and can yield better results compared to grid and random search~\cite{Bergstra_2011}. We use the TPE implementation~\footnote{\url{https://github.com/optuna/optuna}} in the Optuna~\cite{Akiba_2019} library for our pipeline. We train models 
 with different hyperparameter combinations suggested by TPE on $\mathcal{D}_{\train^*}$ and calculate the $\avg$($\Recall$) for $\mathcal{D}_{\val}$. The LSTM model (M9) is trained for $150$ epochs on $\mathcal{D}_{\train^*}$ and stops training if the validation loss did not decrease for $15$ epochs. A model checkpoint with the minimum validation loss is chosen to calculate the $\avg$($\Recall$) on $\mathcal{D}_{\val}$. Finally, the LSTM model with minimum $\avg$($\Recall$) on $\mathcal{D}_{\val}$ over all sampled hyperparameter combinations is chosen for final evaluation. For RF and XGB we use the best set of hyperparameters to train a model on the complete training data $\mathcal{D}_{\train}$. The resulting model is used to calculate predictions on the current test data ($\mathcal{D}_{\test} \gets \mathcal{D}_{t_{\pred}}$) in order to obtain the estimated customer sentiment $\mathbf{\hat{y}}_{t_{\pred}}$. The hyperparameters for the different classifiers are listed in Appendix \ref{sec:hp}.
 For evaluation, we calculate $\Recall$ based on $\mathbf{\hat{y}}_{t_{\pred}}$ and $\mathbf{y}_{t_{\pred}}$ over the whole year ($t_{\pred} \in \{t_0+53, \ldots,t_0+104$\}). This measures the percentage of escalations we would have predicted in one year if we would look at the $N$ largest estimated customer sentiments at each week.
In deployment (Algorithm \ref{alg:deployment}), we provide information regarding the customer sentiment in the current week ($t_{\pred} + 1$) based on $\mathbf{\hat{y}}_{t_{\pred}}$
 since we only have the full data available up to and until $t_{\pred}$. 
 The source code for our experiments is available on GitHub\footnote{\url{https://github.com/annguy/customer-sentiment-monitor}}.
\begin{algorithm}[h!]
    \caption{Build Dataset}
    \label{alg:build_dataset}
    \begin{algorithmic}[1]
        \State  $\mathcal{D} = \emptyset$ \Comment{Dataset} \label{l1}
        \For{all customers $i \in I$} \label{l2}
            \For{$t_{\pred} = t_0:t_0+104$}  \Comment{for every week for 2 years create samples starting from an arbitrary time point $t_0$} \label{l3}
            \If{customer exists since $t_{\pred}-10$}  \label{l4}
                \For{$t = t_{\pred}-10+1:t_{\pred}$} 
                 \Comment{for every week in look-behind window of size $10$} \label{l5}
                    \State extract log features $\rightarrow \mathbf{x}_{i, t, log}$ \label{l6}
                    \State extract enterprise features $\rightarrow \mathbf{x}_{i, t, ent}$ \label{l7}
                \EndFor 
                \State $\mathbf{x}_{i, t_{\pred}} \gets (\mathbf{x}_{i, t, log}, \mathbf{x}_{i, t, ent}), \forall t \in \{t_{\pred}-10+1, \ldots,t_{\pred}\}$  \Comment{feature vector for customer $i$ and $t_{\pred}$} \label{l8}
                \If{$\exists t_{i, \esc} \in T_{i, \esc} \land  t_{i, \esc} \in \{t_{\pred}+1, t_{\pred}+2\}$} \Comment{escalation flag within the next $2$ weeks?} \label{l9}
                    \State $\mathbf{y}_{i, t_{\pred}} \gets 1$ \label{l10}
                \Else  \label{l11}
                    \State $\mathbf{y}_{i, t_{\pred}} \gets 0$ \label{l12}
                \EndIf
                \State  $\mathcal{D} \gets \mathcal{D}_{i, t_{\pred}} = (\mathbf{x}_{i, t_{\pred}}, \mathbf{y}_{i, t_{\pred}})$ \Comment{add sample to dataset} \label{l13}
        \EndIf
            \EndFor
            \For{all escalation flags $t_{i, \esc} \in T_{i, \esc}$} \label{l14}
                \State $\mathcal{D} \gets \mathcal{D} \setminus  \mathcal{D}_{i, t_{i, \esc}:t_{i, \esc}+10+4}$
                \Comment{discard infected intervals after escalation flags from dataset} \label{l15}
            \EndFor
        \EndFor
    \end{algorithmic}
\end{algorithm}

\begin{algorithm}[h!]
    \caption{Training and Validation}
    \label{alg:train_val}
    \begin{algorithmic}[1]
        \For {$t_{\pred} = t_0+53:t_0+104$} \Comment{simulate one year of application} \label{l16}
            \State $\mathcal{D}_{train} \gets \mathcal{D}_{t_{\pred}-53:t_{\pred}-2}$ \Comment{52 weeks training data}
            \State $\mathcal{D}_{train^*} \gets \mathcal{D}_{t_{\pred}-53:t_{\pred}-13}$ \Comment{41 weeks for training for model selection}
            \State $\mathcal{D}_{val} \gets \mathcal{D}_{t_{\pred}-11:t_{\pred}-2}$ \Comment{10 weeks for validation data}
            \State $\mathcal{D}_{test} \gets \mathcal{D}_{t_{\pred}}$ \Comment{current test samples}
            \State model selection using $\mathcal{D}_{train^*}$ and $\mathcal{D}_{val}$ based on $\avg(\Recall)$ \Comment{model selection using TPE sampler}
            \State train model with best hyperparameter on  $\mathcal{D}_{train}$
            \State test model on $\mathcal{D}_{test} \rightarrow \mathbf{\hat{y}}_{t_{\pred}}$ \Comment{estimated customer sentiments}
        \EndFor
        \State Calculate $\Recall$ based on $(\mathbf{\hat{y}}_{t_{\pred}}, \mathbf{y}_{t_{\pred}}), \forall t_{\pred} \in \{t_0+53, \ldots,t_0+104\}$ \label{l24}
    \end{algorithmic}
\end{algorithm}

\begin{algorithm}[h!]
    \caption{In Deployment}
    \label{alg:deployment}
    \begin{algorithmic}[1]
        \State current week is $t_{\pred}+1$ 
        \State output $\mathbf{\hat{y}}_{t_{\pred}}$ and corresponding SHAP values
    \end{algorithmic}
\end{algorithm}

\section{Results}
\label{sec:results}
\begin{figure}[h!]
  \centering
  \includegraphics[width=\linewidth]{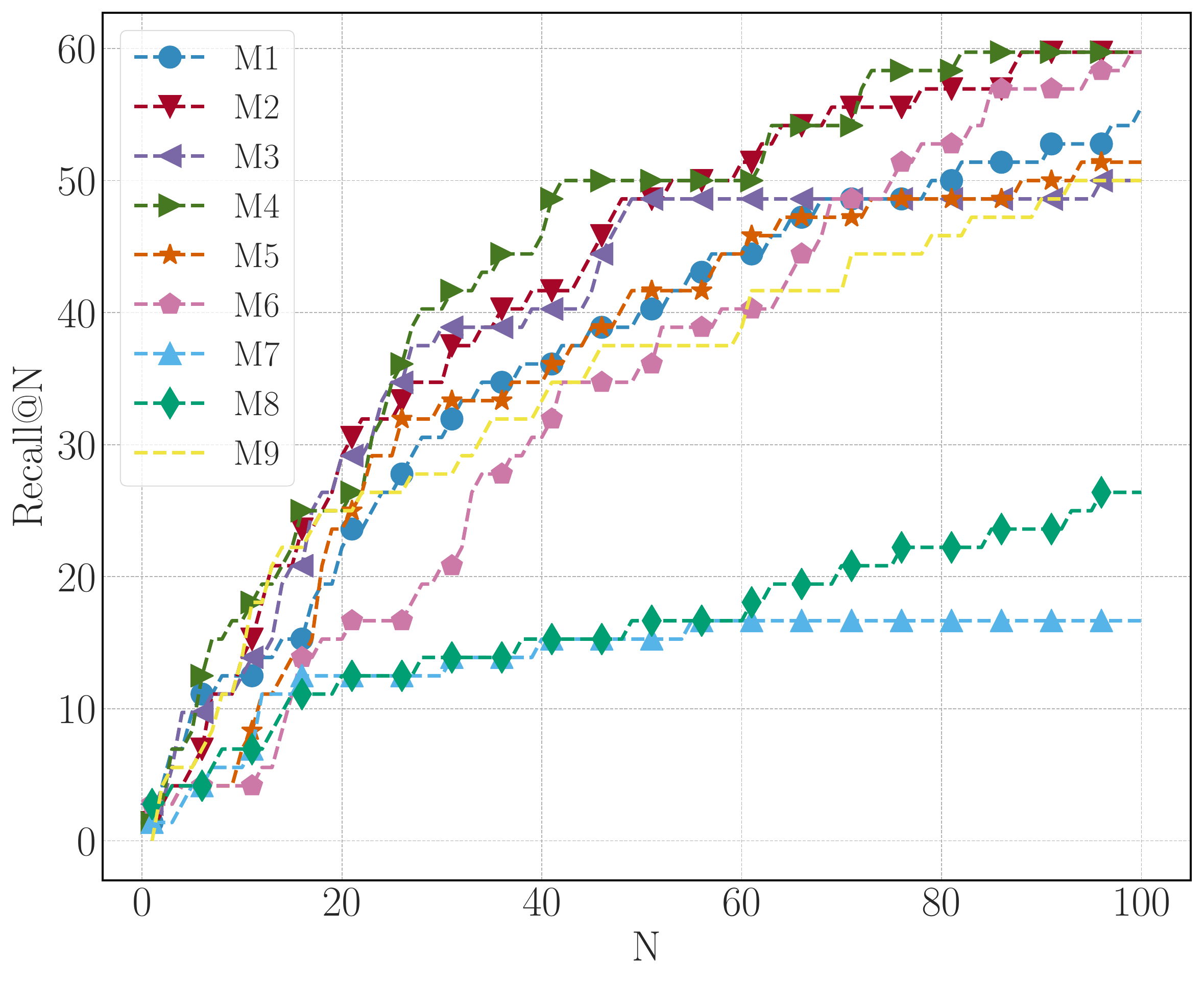}
  \caption{Recall@N curves for N $\in \{1, 2, \ldots, 100\}$ for all model configurations listed in Table \ref{tab:exp}.}
  \label{fig:results}
\end{figure}

\begin{figure}[h!]
  \centering
  \includegraphics[width=0.9\linewidth]{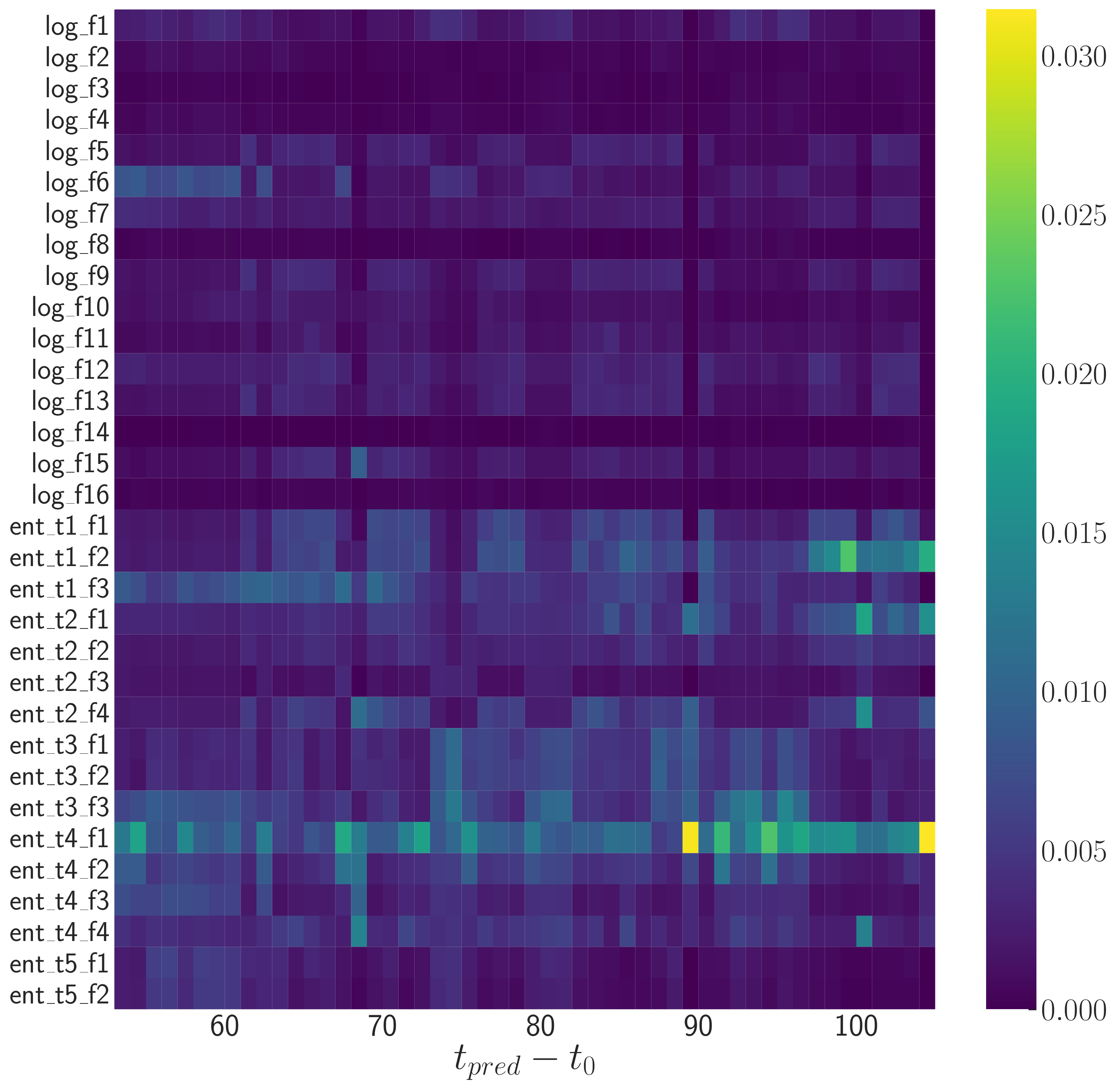}
  \caption{Feature importance for all trained models for configuration M2.}
  \label{fig:feat_importance}
\end{figure}

Figure \ref{fig:results} shows the $\Recall$ values over N $\in  \{1, \ldots, 100\}$ for the different model configurations (Table \ref{tab:exp}). Additionally, Table \ref{tab:results} depicts specific $\Recall$ values for N $\in  \{5, 10, 20, 30, 40, 50, 100\}$ and the overall $\avg(\Recall)$.
\\
\textbf{Early and late fusion:} Comparing M1 vs. M3 and M2 vs. M4 shows that there is a slight overall benefit of late fusion compared to early fusion in terms of $\avg(\Recall)$ ($36.89$ vs. $39.07$ and $42.46$ vs. $44.13$).
\\
\textbf{Feature configuration:} Using log features only (M7 and M8) yielded the worse results with $13.79$ and $16.36$ in terms of $\avg(\Recall)$ respectively. Using enterprise features only (M6 and M7) resulted in $35.99$ and $34.11$ in terms of $\avg(\Recall)$ respectively. Fusing both features yielded consistently better results for all configurations M1-M4 in terms of $\avg(\Recall)$ ($36.89 - 44.13$). Fig. \ref{fig:feat_importance} shows the feature importance of all resulting models for the model configuration M2. One can see that enterprise features are generally more important than log features. Note the small scale and that some log features do have a relatively large feature importance for some weeks. Furthermore, the applied machine learning models can potentially exploit non-linear relationships between the different features.
\\
\textbf{RF and XGB:} RF almost consistently outperformed XGB (M2 vs. M1, M4 vs. M3 and M8 vs. M7) in terms of $\avg(\Recall)$ ($42.46$ vs. $36.89$, $44.13$ vs. $39.07$ and $16.36$ vs. $13.97$). The only exception where XGB performs better than RF is the configuration with enterprise features only (M6 vs. M5) in terms of $\avg(\Recall)$ ($34.11$ vs. $35.99$).
\\
\textbf{Deep Neural Networks}
Our LSTM model (M9) is consistently outperformed by the other models using both feature sets (M1-M4) in terms of $\avg(\Recall)$ ($34.43$ vs. $36.89-44.13$).
\begin{table}[h!]
  \begin{center}
    \begin{tabular}{|c|c|c|c|c|} 
    \hline
     \textbf{Model} &  \textbf{features} & \textbf{classifier} & \textbf{fusion} \\
      \hline
      M1 & $(\mathbf{x}_{log}, \mathbf{x}_{ent})$ & XGB & early\\
      \hline
      M2 & $(\mathbf{x}_{log}, \mathbf{x}_{ent})$ & RF & early \\
      \hline
      M3 & $(\mathbf{x}_{log}, \mathbf{x}_{ent})$ & XGB & late\\
      \hline
      M4 & $(\mathbf{x}_{log}, \mathbf{x}_{ent})$ & RF & late\\
      \hline
      M5 & $\mathbf{x}_{ent}$ & XGB & n.a.\\
      \hline
      M6 & $\mathbf{x}_{ent}$ & RF & n.a.\\
      \hline
      M7 & $\mathbf{x}_{log}$ & XGB & n.a.\\
      \hline
      M8 & $\mathbf{x}_{log}$ & RF & n.a.\\
      \hline
      M9 & $(\mathbf{x}_{log}, \mathbf{x}_{ent})$ & LSTM & early\\
      \hline
    \end{tabular}
        \caption{Different model configurations used for the experiments.}
         \label{tab:exp}
  \end{center}
\end{table}
\begin{table}[h!]
\caption{Numerical results in terms of Recall@N and avg(Recall@N) for all model configurations listed in Table \ref{tab:exp}.}
\label{tab:results}
\begin{adjustbox}{max width=\textwidth}
    \begin{tabular}{|c|c|c|c|c|c|c|c|} 
    \hline
      \textbf{Model} &  \textbf{Recall@10} & \textbf{Recall@20}& \textbf{Recall@30}& \textbf{Recall@40} & \textbf{Recall@50}& \textbf{Recall@100}& \textbf{avg(Recall@N)}\\
      \hline
      M1 & $12.5$ & $22.22$ & $30.56$ & $36.11$ & $40.28$ & $55.56$ & $36.89$\\
      \hline
      M2  & $13.89$ & $\mathbf{29.17}$ &$34.72$ &$41.67$ & $48.61$ & $\mathbf{59.72}$ & $42.46$\\
      \hline
      M3  & $12.5$ & $\mathbf{29.17}$ &$38.89$  &$40.28$ & $48.61$ & $50.0$ & $39.07$\\
      \hline
      M4 & $\mathbf{16.67}$ & $25.0$ & $\mathbf{40.28}$ & $\mathbf{45.83}$ & $\mathbf{50.0}$ & $\mathbf{59.72}$ & $\mathbf{44.13}$\\
      \hline
      M5 & $6.94$ & $23.61$ & $33.33$ & $34.72$ & $41.67$ & $51.39$ & $35.99$\\
      \hline
      M6  & $4.17$ & $15.28$ & $20.83$ & $30.56$ & $36.11$ & $\mathbf{59.72}$ & $34.11$\\
      \hline
      M7 & $5.56$ & $12.5$ & $12.5$  & $15.28$ & $15.28$ & $16.67$ & $13.97$\\
      \hline
      M8 & $6.94$ & $12.5$ & $13.89$ & $15.28$ & $16.67$ & $26.39$ & $16.36$\\
      \hline
      M9 & $13.89$ & $25.0$ & $27.78$ & $33.33$ & $37.5$ & $50.0$ & $34.43$\\
      \hline
    \end{tabular}
 \end{adjustbox}
\end{table}

\section{Discussion}
\label{sec:discussion}
Model configurations M2 and M4 yielded the best results (RF with early and late fusion). Thereby, late fusion (M4) performed slightly better in terms of $\avg(\Recall)$ ($42.46$ vs. $44.13$). In practice, it is harder to compute meaningful SHAP values for the late fusion case since the base classifier are based on different feature sets. Therefore, our practical recommendation is to use M2 with early fusion. RF generally performed better than XGB. We assume that this might be because the gradient-based construction of decision tree ensembles might be more prone to overfitting on the heavily imbalanced dataset with noisy labels. We assume a similar problem with overfitting when using deep learning models for this kind of data. This might explain the inferior performance of M9 despite the potential to better model the temporal structure in the data. 
Furthermore, XGB and LSTM based models are more than 10 times slower to train compared to RF and are harder to tune. We also noticed that the computation of SHAP values \cite{Lundberg_Lee_2017} for LSTM based models is significantly slower compared to XGB and RF.
As a conclusion, we recommend using model configuration M2.
In practice, we estimate that one customer support employee can scan around $10$ customers in depth with our tool per hour. Hence, if for example a team of 5 customer support employees would invest one hour each week using the proposed decision support system with model M2, they could potentially prevent around $48.61\%$ of the escalation in a year. Furthermore, the decision support system can help to learn which specific problems, depicted in the designed features, lead to escalation, as well as to identify customers which have similar problems and might need special attention.
%
In the following we explain in detail how the designed decision support system could be used in practice.
\begin{figure}[h]
  \centering
  \includegraphics[width=0.95\linewidth]{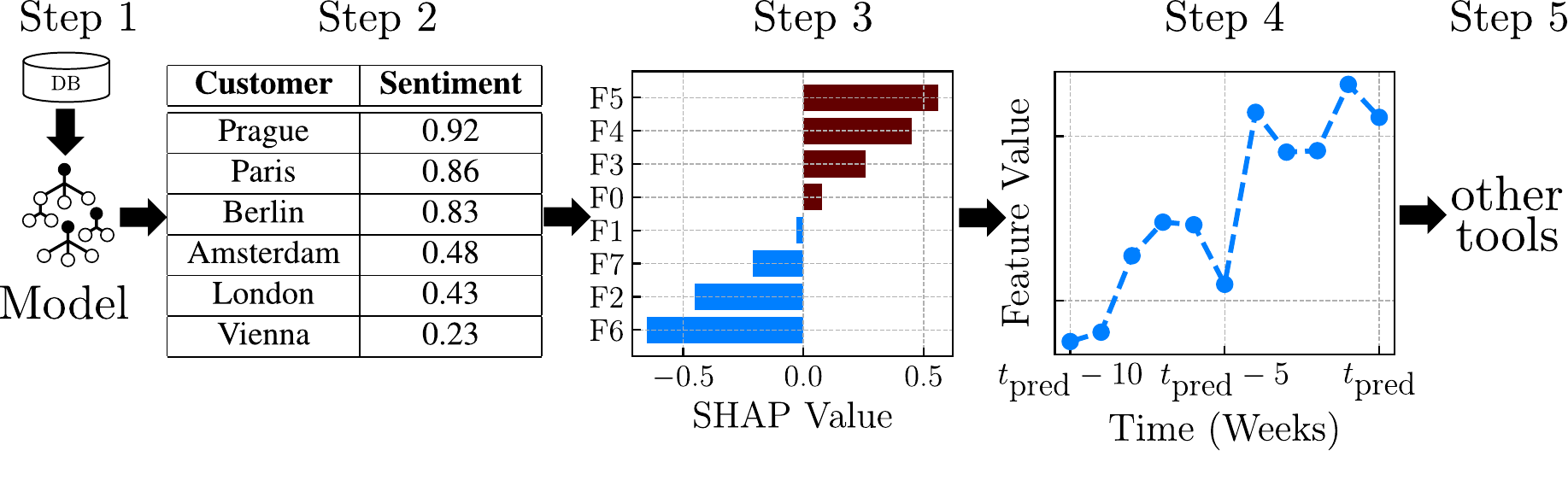}
  \caption{Schematic overview of the envisioned workflow.}
  \label{fig:workflow}
\end{figure}
\subsection{Proposed Workflow}
The following section briefly outlines how the data-driven decision support system is integrated into a productive environment and how it could be used by customer support employees.\\
The envisioned workflow, which is how this system is mainly used, can be grouped into five distinct steps, which are outlined in Fig. \ref{fig:workflow}.
\begin{itemize}
    \item \textbf{Step 1: Producing new predictions} This step is fully automated and all relevant processes are triggered at the beginning of each week. 
    The first process within this step is to load the most recent raw data for all monitored customers from the respective data sources and to conduct the necessary preprocessing steps.
    Afterwards, all available samples up to this point for which labels can be defined are used to train a new model. 
    Once a new model has been trained, it is used to predict the customer sentiment of all monitored customers. 
    Additionally, the SHAP values for each prediction and each feature are calculated.
    Predictions and SHAP values are then copied to a database and automatically loaded into an interactive dashboard which serves as a user interface.
    \item \textbf{Step 2: Identifying high risk customers} One element of the user interface is an interactive table showing the most recent predictions for all monitored customers, along with some additional information about each customer, e.g. location and operated system type.
    Within this step, the user identifies a system within his or her area of responsibility with a particularly high probability for an escalation within the following two weeks.
    Once a customer has been identified, it can be selected, which reduces the information shown on the user interface to just the relevant information about the customer in question.
    \item \textbf{Step 3: Single out the most relevant features with SHAP values} Knowing only which customers are at high risk of causing escalations without knowing why is only of limited use. 
    In order to explain why a specific customer has a high probability for escalation (customer sentiment), SHAP values for each prediction and each feature are displayed in the user interface.
    With such a visualization, the user can easily single out one or a few features which, according to their respective SHAP values, have a large effect on the customers sentiment. 
    By selecting these specific features, the information shown on the interface is further reduced, now only displaying information connected to the selected customer and the selected feature or features.
    \item \textbf{Step 4: Analyzing specific features} Once a set of few features has been selected, the user is provided with the actual values of these features and how these values have been changed over the past weeks.
    With this information, the user can easily identify open, yet unresolved, tickets and see immediately for how long a specific ticket has been unresolved. 
    Another example could be the accumulation of specific malfunctions reflected in log features.
    \item \textbf{Step 5: In depth analysis of certain problem} At this point, the experienced user probably has a good idea of where a potential problem with the customer in question might be found (e.g. unresolved tickets, spare parts, software issues).
    For an in depth analysis of  ticket data or consumed spare parts, other tools are already available which are tailored for such tasks. 
    Therefore, once the user has identified the potential root cause for a bad customer sentiment, he or she is provided with a direct link to these external tools in order to continue the analysis as efficient as possible with the goal to act before an escalation occurs.
\end{itemize}
The main idea of a productive use of a data-driven decision support system is to help customer support employees decide which customers to focus on and where to look.
\section{Conclusion and Future Work}
\label{sec:conclusion}
In this paper, we propose a general framework and an interactive workflow with a decision support system. Additionally, we provide a publicly available industrial benchmark dataset, including all code necessary to reproduce or to improve our results. Our designed and implemented decision support system is currently deployed to monitor the customer sentiment of thousands of customers of high-end medical devices worldwide. The explainability of the system helps a variety of end users to identify problems in the field. We demonstrate that using both log and enterprise data-based features enables more effective troubleshooting compared to using either of these data sources alone. Furthermore, the gained insights can help to achieve better and more proactive customer relations, as well as improve product management by focusing on problems which affect the customer the most.
There are some open challenges which could be addressed in future research using the provided benchmark dataset and evaluation framework. For example, more efficient methods for merging log and enterprise data information which preserve explainability can be investigated. Another challenge is to design models that, within the implemented framework, allow to increase its predictive power without trading interpretability. Finally, alternative learning problem formulations, like anomaly detection, could be explored for the task. This could help with the heavy class imbalance present in the benchmark dataset.
\bibliography{main.bib}
\section*{Acknowledgement}
Bjoern Eskofier gratefully acknowledges the support of the German Research Foundation (DFG) within the framework of the Heisenberg professorship programme (grant number ES 434/8-1). We would like to thank Gilles Le Texier, Martin Rothgaengel, Birgi Tamersoy, Mirko Appel and Marie Mecking from Siemens Healthineers for their valuable inputs and discussions. We gratefully acknowledge the support of Siemens Healthineers for this study and for providing the data. We would also like to thank Erick Axxe from the Ohio State University for proofreading our manuscript as a native speaker. 
\appendix
\section{Hyper Parameter}
\label{sec:hp}
Table \ref{tab:hp} summarizes the hyperparamter search space for the classifiers used our experimental study. The sampling strategies are according to the Optuna \cite{Akiba_2019} \footnote{\url{https://github.com/optuna/optuna}} libary. The parameter names correspond to the respective implementation of the classifiers.
\begin{table*}[hbt!]
  \begin{center}
    \begin{tabular}{|c|c|c|c|c|} 
    \hline
     \textbf{classifier} &  \textbf{parameter} & \textbf{range} & \textbf{sampling strategy} \\
      \hline
      \multirow{6}{4em}{RF\\ (M2, M4, M6, M8)} & max\_samples  & \{0.1, 1\}  & uniform \\
      & max\_depth & $\{1, 50\}$ & int \\
      & n\_estimator & $\{100, 1000\}$ & int \\
      & criterion & \{gini, entropy\}& categorical \\
      & ratio & $\{0.1, 1.0\}$ & uniform\\
      & sampling\_strategy & \{over, SMOTE, under\} & categorical \\
      \hline
     \multirow{8}{4em}{XGB (M1, M3, M5, M7)} & learning rate  & $\{1\mathrm{e}{-3}, 1\}$  & loguniform \\
     & max\_depth & $\{1, 50\}$ & int \\
      & n\_estimator & $\{100, 1000\}$ & int \\
      & colsample\_bytree & $\{0.5, 1\}$ & uniform \\
      & subsample & $\{0.5, 1\}$ & uniform \\
      & reg\_lambda & $\{0.1, 10\}$ & loguniform \\
      & ratio & $\{0.1, 1.0\}$ & uniform\\
      & sampling strategy & \{over, SMOTE, under\} & categorical \\
      \hline
      \multirow{11}{4em}{LSTM (M9)} & batch\_size  & $\{32, 64, 128\}$  & categorical \\
      & num\_epochs & $\{150\}$ & fixed \\
      & early\_stopping & $\{15\}$ & fixed \\
      & hidden\_dim & $\{16, 128\}$ & int \\
      & learning\_rate & $\{1\mathrm{e}{-5}, 1\}$ & loguniform \\
      & weight\_decay & $\{0, 0.75\}$ & uniform\\
      & dropout\_prob & $\{0, 0.75\}$ & uniform\\
      & lstm\_layer & $\{1, 2\}$ & int\\
      & lstm\_bidirectional & \{True, False\}  & categorical \\
      & ratio & $\{0.1, 1.0\}$ & uniform\\
      & sampling strategy & \{over, under\} & categorical \\
      \hline
    \end{tabular}
        \caption{Settings for hyperparamter tuning.}
         \label{tab:hp}
  \end{center}
\end{table*}
\end{document}